\documentclass{article}

\PassOptionsToPackage{numbers, sort&compress}{natbib}

\usepackage[preprint]{nips_2018}

\usepackage[utf8]{inputenc}
\usepackage[T1]{fontenc}
\usepackage[british]{babel}

\usepackage{color,xcolor}

\definecolor{dark-red}{rgb}{0.8, 0.0, 0.1803921568627451}
\definecolor{dark-blue}{rgb}{0.0, 0.0, 0.803921568627451}
\definecolor{dark-green}{rgb}{0.0, 0.39215686274509803, 0.0}
\definecolor{dark-orange}{rgb}{0.8, 0.4, 0.0}

\usepackage{amsmath,dsfont,mathtools}
\usepackage{graphicx}

\usepackage{xfrac}

\usepackage{subcaption}

\usepackage{multirow,booktabs,tabularx,float}
\PassOptionsToPackage{hyphens}{url}\usepackage[pdftex,hidelinks,linkcolor={dark-blue}, citecolor={dark-green}, urlcolor={dark-red},bookmarksopen,linktoc=all]{hyperref}
\usepackage{enumitem} 
\usepackage[stretch=15,shrink=15]{microtype} 
\usepackage{xspace} 

\newcommand{\diff}{\mathrm{d}}

\newcommand{\ie}{{i.\,e.}~}

\newcommand{\carl}{\textsc{Carl}\xspace}
\newcommand{\rolr}{\textsc{Rolr}\xspace}
\newcommand{\sally}{\textsc{Sally}\xspace}
\newcommand{\sallino}{\textsc{Sallino}\xspace}
\newcommand{\cascal}{\textsc{Cascal}\xspace}
\newcommand{\rascal}{\textsc{Rascal}\xspace}

\newcommand{\alice}{\textsc{Alice}\xspace}
\newcommand{\alices}{\textsc{Alices}\xspace}

\newcommand{\intx}{\int \! \diff x\;}

\newcommand{\intz}{\int \! \diff z\;}

\newcommand{\intxz}{\int \! \diff x \, \diff z\;}

\newlength{\hhatheight}

\DeclareMathOperator*{\argmin}{arg\,min}

\newcolumntype{R}{>{\raggedleft\arraybackslash}X}%
\newcolumntype{L}{>{\raggedright\arraybackslash}X}%

\setlist[itemize]{itemsep=1pt,parsep=1pt, topsep=1pt}

\bibpunct{[}{]}{,}{n}{,}{,}

\addto\captionsbritish{}

\title{Likelihood-free inference with an improved cross-entropy estimator}

\author{Markus Stoye,$^1$ Johann Brehmer,$^2$ Gilles Louppe,$^3$ Juan Pavez,$^4$ and Kyle Cranmer$^2$ \\
{}$^1$\, Department of Physics and Data Science Institute, Imperial College London \\{}$^2$\, Center for Cosmology and Particle Physics and Center for Data Science, New York University \\{}$^3$\, Department of Electrical Engineering and Computer Science, University of Li\`{e}ge,\\
 {}$^4$\,Federico Santa Mar\'ia Technical University\\
\texttt{markus.stoye@cern.ch}, \texttt{johann.brehmer@nyu.edu}, \texttt{g.louppe@uliege.be},\\
\texttt{juan.pavezs@alumnos.usm.cl}, \texttt{kyle.cranmer@nyu.edu} \\
}
\date{\today}

\begin{document}

\maketitle

\begin{abstract}
We extend recent work (Brehmer, et. al., 2018)  that use neural networks as surrogate models for likelihood-free inference. As in the previous work, we exploit the fact that the joint likelihood ratio and joint score, conditioned on both observed and latent variables, can often be extracted from an implicit generative model or simulator to augment the training data for these surrogate models. We show how this augmented training data can be used to provide a new cross-entropy estimator, which provides improved sample efficiency compared to previous loss functions exploiting this augmented training data. 
\end{abstract}


\section{Introduction}
\label{sec:intro}

Many real-world phenomena are best described by computer simulations. Such simulators often implement a stochastic generative process, which is based on a mechanistic model and parametrized by $\theta$. In practice, these simulators are used to generate samples of observations $x \sim p(x | \theta)$, but the density is only defined implicitly through the simulation code. Often, the generative process involves latent variables and the density
\begin{equation}
  p(x | \theta) = \intz p(x,z | \theta)
\end{equation}
is intractable because of the integral over a large (and possibly highly structured) latent space. 
Without a tractable likelihood, statistical inference on the parameters $\theta$ given observed data $x$ is challenging. This problem has prompted the development of \emph{likelihood-free inference} methods such as Approximate Bayesian Computation~\cite{rubin1984, beaumont2002approximate, Alsing:2018eau, Charnock:2018ogm} and neural density or neural density ratio estimation algorithms~\cite{kanamori2009least, 2012arXiv1212.1479F, 2014arXiv1410.8516D, 2015arXiv150505770J, Cranmer:2015bka, Cranmer:2016lzt, 2016arXiv160508803D, NIPS2016_6084, 2016arXiv160206701P, 2016arXiv161110242D, 2016arXiv160502226U, 2016arXiv160903499V, 2016arXiv160605328V, 2016arXiv160106759V, gutmann2017likelihood, 2017arXiv170208896T, 2017arXiv170707113L, 2017arXiv170507057P, 2018arXiv180507226P}. Nearly all of these established methods treat the simulator as a black box and only use its capability to generate samples for a specified values of $\theta$.

In Refs.~\cite{prl, prd, nips} a new paradigm was introduced that exploits additional information that can be extracted from the simulation. In particular, within the simulation where the latent variables $z$ are available, it is often possible to extract the \emph{joint likelihood ratio}
\begin{equation}
  r(x,z | \theta_0, \theta_1) = \frac{p(x,z | \theta_0)}{p(x,z | \theta_1)}
\end{equation} 
and the \emph{joint score}
\begin{equation}
  t(x,z | \theta_0) = \nabla_{\theta}  \log p(x,z | \theta)  \Biggr|_{\theta_0} \,,
\end{equation} 
which are conditioned on the latent variables $z$ corresponding to a particular sample.

It was then shown that certain loss functionals $L[g(x)]$, which depend on the joint likelihood ratio and the joint score, are minimized by the likelihood ratio
\begin{equation}
  g^*(x) \equiv \argmin_{g(x)} L[g(x)] = r(x | \theta_0, \theta_1) \equiv \frac{p(x | \theta_0)} {p(x | \theta_1)} \,,
\end{equation}
an otherwise intractable quantity. This motivates a family of new techniques for likelihood-free inference in which the the joint likelihood ratio and joint score are used as training data for neural networks. These networks serve as surrogate models for the intractable likelihood or likelihood ratio. Experiments showed these new methods to be more sample-efficient than previously established neural density and neural density ratio estimation techniques. The authors of Refs.~\cite{prl, prd, nips} coined the term ``mining gold'' for the process of extracting the joint likelihood ratio and joint score from the simulator -- while the augmented data require some effort to extract, they are extremely valuable. 

While the loss functionals originally proposed in Refs.~\cite{prl, prd, nips} have the correct minima, they are not necessarily the most sample efficient. In particular, the proposed mean squared error (MSE) losses are often dominated by few samples with large joint likelihood ratios. 
Here we extend and improve that original work with two new algorithms for likelihood-free inference. The key improvement are new loss functions, which use an improved estimator for the cross entropy based on the joint likelihood ratio and joint score.

After introducing these new algorithm in Sec.~\ref{sec:method}, we show its performance in a problem from particle physics in Sec.~\ref{sec:experiments}, before giving our conclusions in Sec.~\ref{sec:conclusions}.

\section{Cross-entropy estimation with augmented data}
\label{sec:method}

Consider the problem of estimating the likelihood ratio $r(x |  \theta_0, \theta_1)$ based on samples $(x_i, z_i) \sim p(x, z | \theta_0)$, labeled with $y_i=0$; samples $(x_i, z_i) \sim p(x,z | \theta_1)$, labeled $y_i=1$; and the joint likelihood ratio $r(x_i, z_i | \theta_0, \theta_1)$ and joint score $t(x_i, z_i | \theta_0)$.

The familiar binary cross-entropy loss functional is defined as
\begin{equation}
  L[\hat{s}(x)] =  - \intx  \Bigl[ p(x|y=1) \, \log (\hat{s}(x))
  + p(x|y=0) \, \log (1 - \hat{s}(x)) \Bigr] \;.
\label{eq:xe_normal}
\end{equation}
For balanced samples ($p(\theta_0) = p(\theta_1) = \sfrac{1}{2}$) we have 
\begin{align}
  p(x,z) &= \frac {p(x,z | \theta_0) + p(x,z | \theta_1)} 2 \\
  s(x,z | \theta_0, \theta_1) &= p(y = 1 | x, z) = \frac 1 {r(x, z | \theta_0, \theta_1) + 1} = \frac {p(x, z | \theta_1)} {p(x, z | \theta_0) + p(x, z | \theta_1)}  \;, 
\end{align}
which allows us to rewrite Eq.~\ref{eq:xe_normal} as
\begin{equation}
  L[\hat{s}(x)] =  - \intxz  p(x,z) \Bigl[ s(x,z | \theta_0, \theta_1) \, \log (\hat{s}(x))
  + (1 - s(x,z | \theta_0, \theta_1)) \, \log (1 - \hat{s}(x)) \Bigr]\,.
\end{equation}
It is straightforward to show that this loss functional is minimized by
\begin{equation}
  s^*(x) \equiv \argmin_{\hat{s}(x)} L[\hat{s}(x)] = s(x | \theta_0, \theta_1) = p(y = 1 | x) = \frac 1 {r(x | \theta_0, \theta_1) + 1} = \frac {p(x | \theta_1)} {p(x | \theta_0) + p(x | \theta_1)}  \,.
  \label{eq:xe_minimum}
\end{equation}

To use the cross entropy to train a surrogate model with a finite number of samples, we need a tractable estimator for the cross entropy. The standard estimator, as used for instance in the \carl inference method~\cite{Cranmer:2015bka}, is given by
\begin{equation}
  L_{\carl} [\hat{s}(x)] = - \frac 1 N \sum_{(x_i, y_i)} \Bigl[ y_i \log (\hat{s}(x_i)) + (1 - y_i) \log (1 - \hat{s}(x_i)) \Bigr] \,.
  \label{eq:standard_xe}
\end{equation}
The $y_i \in \{0,1\}$ act as an unbiased, but high-variance estimator of $s(x_i,z_i | \theta_0, \theta_1)$.  In the limit of infinite samples, this estimator therefore has the correct minimum of Eq.~\eqref{eq:xe_minimum}, but for finite sample sizes it may suffer from high variance.

With the availability of the joint likelihood ratio $r(x_i,z_i | \theta_0, \theta_1)$ from the simulator, the $s(x_i, z_i | \theta_0, \theta_1)$ are tractable and we can define the alternative estimator
\begin{equation}
  L_{\alice} [ \hat{s}(x) ] = - \frac 1 N \sum_{(x_i,z_i) \sim p(x_i,z_i)} \Bigl[ s(x_i,z_i | \theta_0, \theta_1) \log (\hat{s}(x_i)) + (1 - s(x_i,z_i | \theta_0, \theta_1)) \log (1 - \hat{s}(x_i)) \Bigr] \,.
  \label{eq:improved_xe}
\end{equation}
By using the exact $s(x,z | \theta_0, \theta_1)$ rather than the $y_i \in \{0,1\}$, the samples drawn according to $y=0$ also provide information about the second $y=1$ term in the loss function, and vice versa. By minimizing the loss function we get an estimator $\hat{s}(x)$ and thus a likelihood ratio estimator
\begin{equation}
  \hat{r}(x | \theta_0, \theta_1) = \frac {1 - \hat{s}(x)} {\hat{s}(x)} \,.
\end{equation}
This defines the \alice inference method\footnote{\textbf{A}pproximate \textbf{l}ikelihood with \textbf{i}mproved \textbf{c}ross-entropy \textbf{e}stimator}, which consists of mining the joint likelihood ratio from the simulator, training a neural network on the improved cross-entropy estimator in Eq.~\eqref{eq:improved_xe}, and using this surrogate model for statistical inference on $\theta$.


It is to be expected that a likelihood ratio estimator based on the \alice estimator for the cross-entropy should outperform the \carl method, which is based on the standard cross-entropy estimator in Eq.~\eqref{eq:standard_xe}. The more interesting question is how it stacks up against the \rolr technique introduced in Refs.~\cite{prl, prd, nips}, in which the loss function
\begin{multline}
  L_{\rolr} [\hat{r}(x)] = \frac 1 N \sum_{(x_i,y_i, z_i) } \Bigl[ y_i \, |r(x_i,z_i | \theta_0, \theta_1) - \hat{r}(x| \theta_0, \theta_1)|^2 \\
    + (1-y_i)  \, \left |\frac 1 {r(x_i,z_i | \theta_0, \theta_1)} - \frac 1 {\hat{r}(x| \theta_0, \theta_1)} \right|^2 \Bigr]
    \label{eq:rolr_loss}
\end{multline}
is minimized. In the limit of infinite samples it is minimized by $r(x| \theta_0, \theta_1)$.  But here each event only contributes to either the squared error on $r$ or on $1/r$ term, which might lead to a higher variance.

In analogy to the \cascal and \rascal methods of Refs.~\cite{prl, prd, nips}, we can define an additional inference method which uses the joint score, \ie an additional piece of information that describes the local (tangential) behavior of the likelihood function. If a parameterized likelihood ratio estimator is implemented with a differentiable architecture such as a neural network, we can calculate the gradient of the output $\hat{s}(x | \theta_0, \theta_1)$ with respect to $\theta_0$ and similarly calculate the corresponding score
\begin{equation}
  \hat{t}(x | \theta_0, \theta_1) = \nabla_\theta \log \hat{r}(x | \theta_0, \theta_1) = \nabla_\theta \log \left(\frac{1 - \hat{s}(x_i |  \theta, \theta_1)}{\hat{s}(x_i |  \theta, \theta_1)}  \right)
\end{equation}
of the $\hat{r}$ estimator. For a perfect $\hat{r}$ (or equivalently $\hat{s}$) estimator, this corresponding score $\hat{t}$ will also minimize the squared error loss with respect to the joint score $t(x,z | \theta_0, \theta_1)$, which can be extracted from the simulator~\cite{prl, prd, nips}. Turning this argument around, we can use the joint score to guide the training of the estimator. This is the idea behind the \alices\footnote{\textbf{A}pproximate \textbf{l}ikelihood with \textbf{i}mproved \textbf{c}ross-entropy \textbf{e}stimator  and \textbf{s}core} technique, which is based on the loss function
\begin{multline}
  L_{\alices} [ \hat{s}(x | \theta_0, \theta_1) ]
= - \frac 1 N \sum_{(x_i,z_i) \sim p(x_i,z_i)} \Biggl[ s(x_i,z_i | \theta_0, \theta_1) \log (\hat{s}(x_i)) \\
+ (1 - s(x_i,z_i | \theta_0, \theta_1) \log (1 - \hat{s}(x_i)) \\
+ \alpha \, (1 - y_i)
\left| t(x_i,z_i | \theta_0, \theta_1) - \nabla_\theta \log \left(\frac{1 - \hat{s}(x_i |  \theta, \theta_1)}{\hat{s}(x_i |  \theta, \theta_1)}  \right) \Biggr|_{\theta_0} \right|^2 \Biggr] \,.
\end{multline}
The factor $(1 - y_i)$ is necessary to guarantee the correct minimum of the squared error on the score. The hyper-parameter $\alpha$ weights the two terms in the loss function. This loss is the natural extension of the the \cascal loss function, but we expect it to reduce the variance compared to the \cascal approach for finite sample size. An interesting question is how it performs compared to the \rascal approach, which similarly augments the \rolr loss in Eq.~\eqref{eq:rolr_loss} with the score term.

\section{Experiments}
\label{sec:experiments}

We experiment with the new methods in the particle physics problem introduced in Refs.~\cite{prl,prd}. In  this real-world problem, the outcome of proton-proton collisions is characterized by 42 observables, from which likelihood ratios and confidence limits on two model parameters are derived.  We first consider an idealized setting neglecting the detector response where the likelihood function is tractable, which provides us with ground truth that can be used to evaluate the performance of the algorithms. For a detailed description of the setup, see Ref.~\cite{prd}.

In addition to the \carl, \rolr, \cascal,  and \rascal techniques described above, we also compare to the \sally and \sallino methods. \sally and \sallino approximate a statistical model that is accurate in the neighborhood of $\theta=(0,0)^T$. The methods are very sample efficient, but make approximations that limit their asymptotic performance.

Except for the new loss functions, we used the same architectures and hyper-parameters as in Ref.~\cite{prd}. In particular, we use fully connected networks with five hidden layers, 100 units each, and $\tanh$ activation functions for both approaches. For \alices we use $\alpha = 5$, which was found to give a good performance for the closely related \cascal method~\cite{prd}. 

\begin{table}
\centering
  \begin{tabular}{l rrr}
    \toprule
    \multirow{2}{*}{Strategy} & \multicolumn{3}{c}{Expected MSE} \\
    \cmidrule{2-4}
    & $10^4$ training samples & $10^5$ training samples & $10^7$ training samples \\
    \midrule
   Histogram &  &  & $0.0561$\\
   \carl & $\mathbf{0.1743}$ & $\mathbf{0.1672}$ & $\mathbf{0.0124}$\\
   \midrule
   \rolr & $0.1345$ & $0.0396$ & $0.0032$\\
   \cascal & $0.1715$ & $0.1652$ & $0.0008$\\
   \rascal & $0.0449$ & $0.0100$ & $0.0009$\\
   \alice & $0.0510$ & $\mathbf{0.0076}$ & $\mathbf{0.0004}$\\
   \alices & $\mathbf{0.0339}$ & $0.0111$ & $0.0013$\\
   \midrule
   \sally & $\mathbf{0.0261}$ & $\mathbf{0.0146}$ & $\mathbf{0.0132}$\\
   \sallino & $0.0319$ & $0.0227$ & $0.0213$\\
    \bottomrule\\
  \end{tabular}
    \caption{Fidelity of different strategies in  the idealized scenario, using  training sets of various sizes. We use the expected mean squared error as defined in Ref.~\cite{prd} as a performance metric. The new methods, \alice and \alices, outperform  \rascal, \cascal, and \rolr. The \sally and \sallino methods are very sample efficient, but make approximations that limit their asymptotic performance.}
  \label{tbl:results}
\end{table}

\begin{figure}
\centering
  \includegraphics[width=0.49\textwidth]{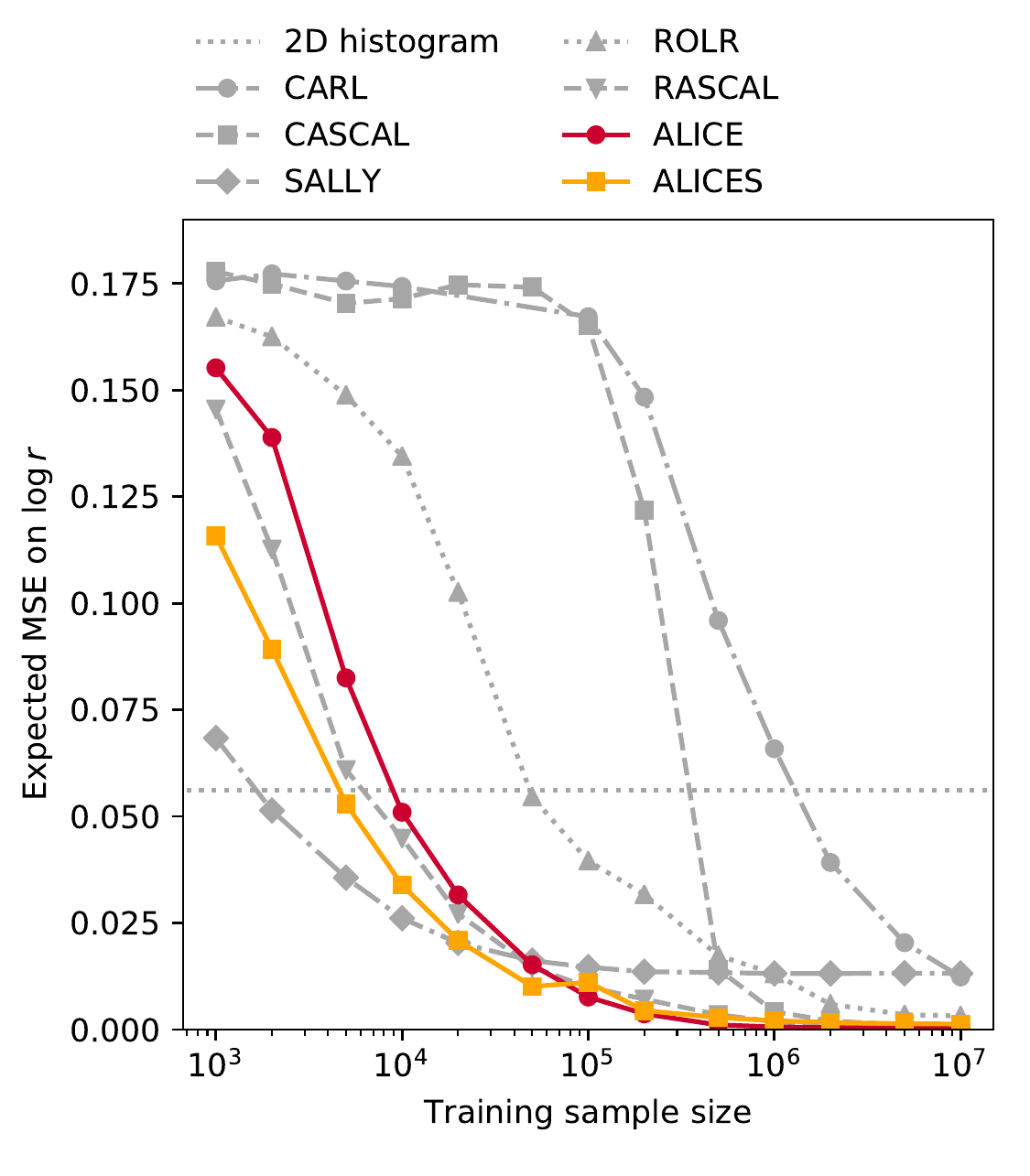}%
  \caption{Estimator fidelity in the idealized scenario as a function of the training sample size. As a metric we use the expected mean squared error on the log likelihood ratio, see Ref.~\cite{prd}.  The new methods are more sample efficient than the similar \rolr and \rascal techniques.}
  \label{fig:sample_size}
\end{figure}

Table~\ref{tbl:results} and Fig.~\ref{fig:sample_size} show the quality of the likelihood ratio estimate based on various sized training samples for the new methods and compares them to the inference techniques presented in Ref.~\cite{prd}. As a performance metric we use an expected mean squared error on the log likelihood ratio, as defined in Ref.~\cite{prd}. 

Unsurprisingly, the \alice and \rolr methods clearly outperform \carl, which does not have access to the joint likelihood ratio. More significantly, we find that \alice outperforms and \rolr, which does have access to the joint likelihood ratio. 
We conjecture that this improvement can be attributed to the lower variance of the cross-entropy compared to the squared error.
More surprisingly, the \alice method also outperforms the \rascal method for larger training sample sizes ($\ge 10^5$), even though \alice does not have access to the joint score.

For smaller training sample sizes ($\le 10^5$) the \alices method outperforms the \alice method, which is not surprising given the additional information available during training. For larger training sample sizes ($\ge 10^5$), the variance of the score actually deteriorates the performance of \alices compared to \alice. We did not perform hyper-parameter tuning for $\alpha$ as a function of the training sample size, which should ensure that \alices performs at least as well as \alice. 
We leave a systematic tuning of the $\alpha$ parameter and an analysis of sources of variance in this approach for future work.

\begin{figure}
  \centering 
  \includegraphics[width=0.49\textwidth]{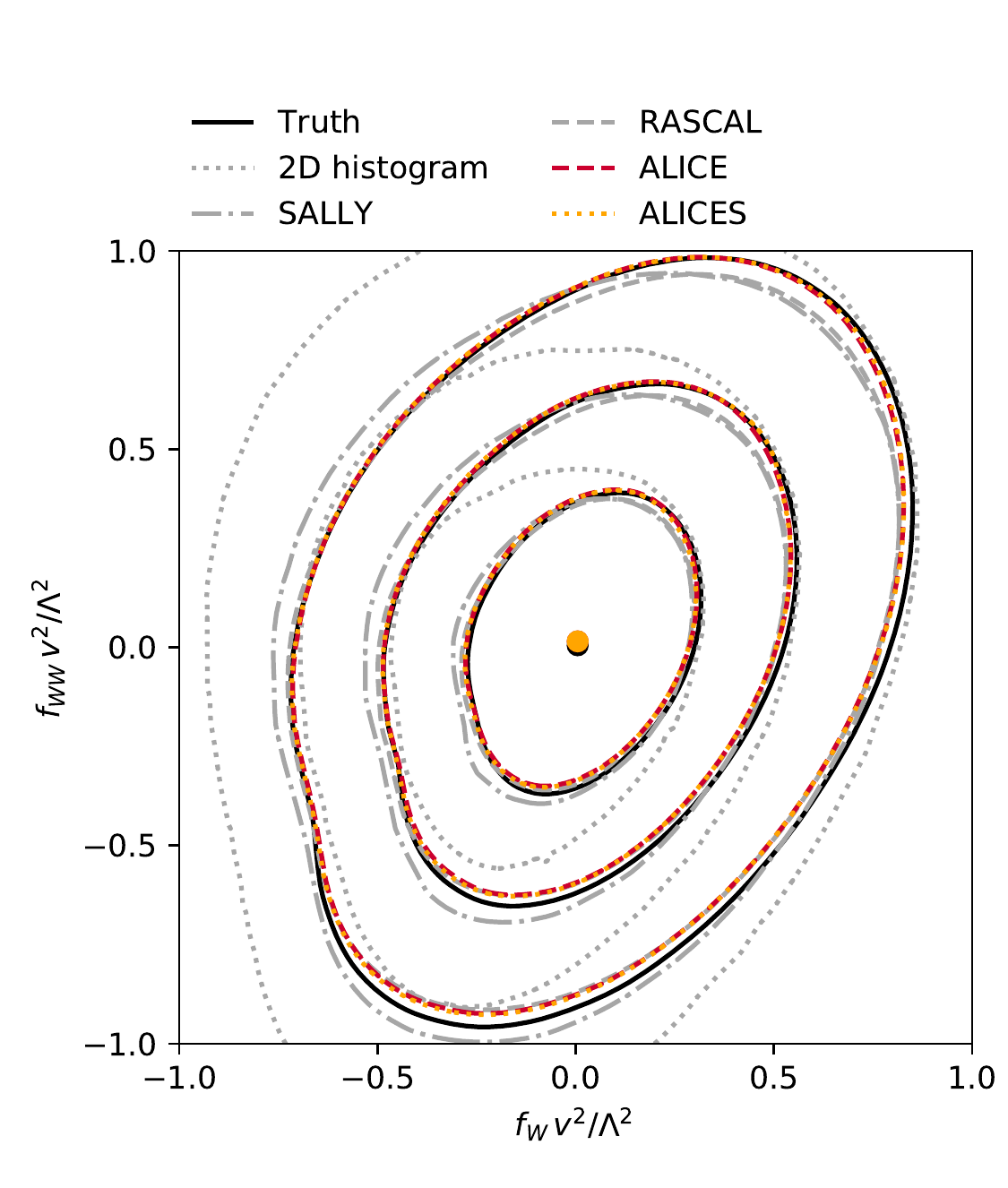}%
  \includegraphics[width=0.49\textwidth]{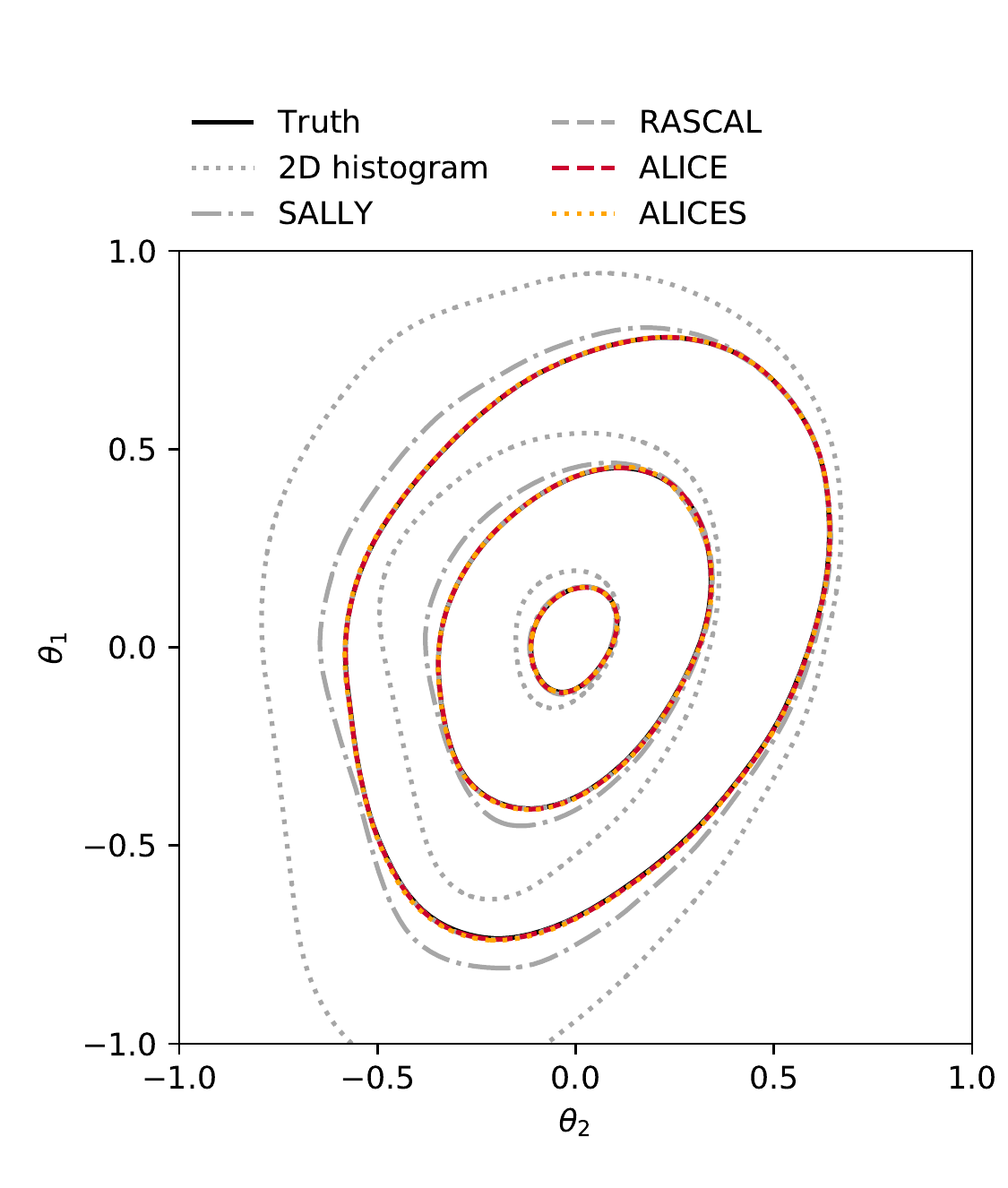}%
  \caption{Expected exclusion limits on the model parameters in the idealized scenario for different inference methods. We assume 36 events distributed according to $\theta = (0,0)^T$. All estimators are trained on a large data set with $10^7$ samples. Left: construction of exclusion limits based on asymptotic properties of the likelihood ratio. With this method, inefficient estimators can predict overly optimistic exclusion limits, as can be seen for instance for the \rascal method. The new \alice and \alices approaches are less prone to this issue. Right: construction of exclusion limits calibrated with toy experiments (\ie the Neyman construction). In this approach, the intervals will always cover, but might not be optimal. We find an excellent performance of the \alice and \alices methods, virtually indistinguishable from  the \rascal method and the true likelihood ratio.}
  \label{fig:contours}
\end{figure}

Figure~\ref{fig:contours} shows expected exclusion contours at different confidence levels on the two parameters, assuming 36 observed events distributed according to $\theta = (0,0)^T$. The methods are trained on the full training samples of $10^7$ samples. The left panel shows contours constructed based on asymptotic properties of the profile likelihood ratio test statistic. While methods such as \rascal are generally very accurate, with this construction they can sometimes lead to overly optimistic exclusion contours, visible as tighter bounds than the ``truth'' contour. We find that switching to \alice and \alices reduces this issue, but does not entirely solve it.

The right panel of Fig.~\ref{fig:contours} shows exclusion contours based on the frequentist confidence intervals calibrated with toy experiments. This Neyman construction guarantees coverage: while the limits from any approach may be worse than the optimal limits, they will never be overly optimistic. As test statistics we use the likelihood ratio with respect to the $\theta = (0,0)^T$, which explains why the contours are generally stronger than in the left panel. We find that both \alice and \alices, like \rascal and \cascal of Refs.~\cite{prl, prd, nips}, lead to limits that are virtually indistinguishable from the ideal limits based on the true likelihood ratio. 

\begin{figure}
  \centering
  \includegraphics[width=0.49\textwidth]{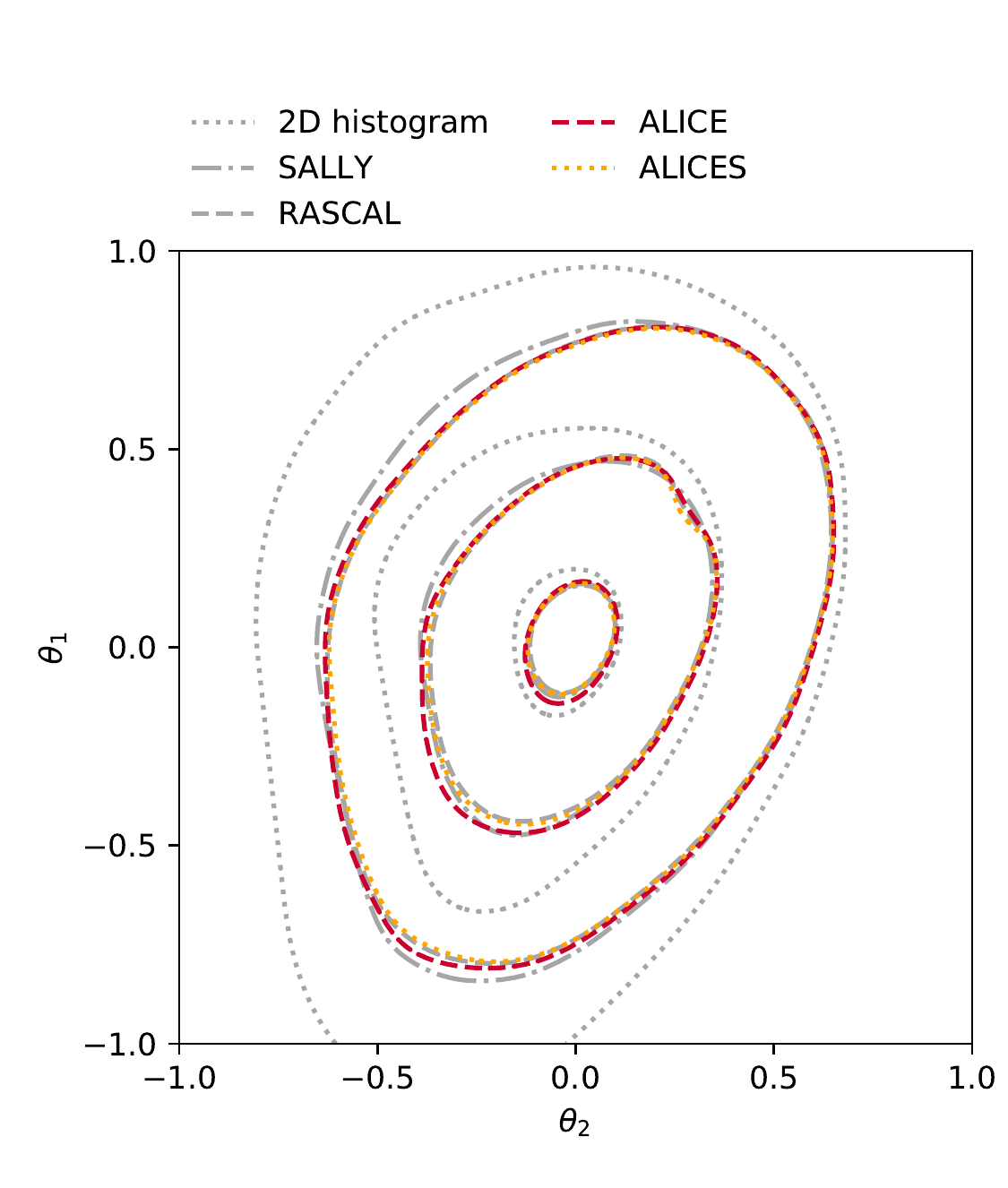}%
  \includegraphics[width=0.49\textwidth]{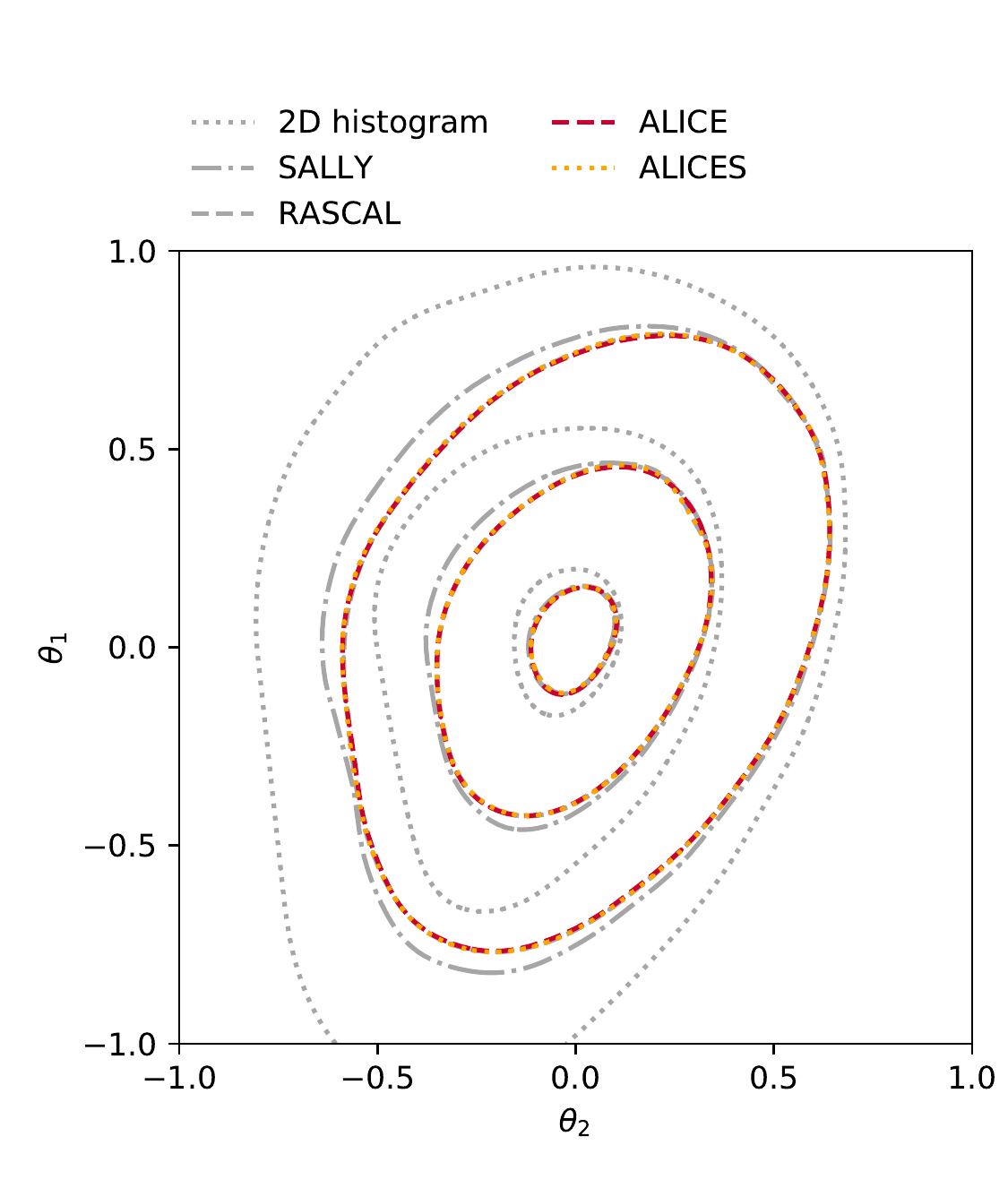}%
  \caption{Expected exclusion limits on the model parameters in the scenario with detector effects for different inference methods. We construct the contours with the Neyman construction, which guarantees the confidence intervals will cover. The intervals are based on 36 events distributed according to $\theta = (0,0)^T$. The  estimators are trained on data sets with $10^5$ (left) or $10^7$ (right) samples. The \alices method leads to strong limits, comparable to the \rascal technique.}
  \label{fig:contours_smearing}
\end{figure}

Finally, in Fig.~\ref{fig:contours_smearing} we show similar expected exclusion contours, but in a more realistic setup in which the parton shower and detector effects are described with approximate smearing functions, which makes the true likelihood intractable. In this situation, we cannot compare the likelihood ratio estimators to the ground truth. Instead, we show the expected contours based on the Neyman construction, similar to the right panel of Fig.~\ref{fig:contours}. In the left panel we show results for limited training samples of only $10^5$ events. In this setup, \alices allows for strong limits, comparable to \rascal and slightly better than for \alice. The right panel demonstrates that with the full training sample the results of \rascal, \alice, and \alices are indistinguishable.



\section{Conclusions}
\label{sec:conclusions}

In this work, we have extended recently developed inference techniques for the setting in which the likelihood is only implicitly defined through a stochastic generative model or simulator. By exploiting the joint likelihood ratio that can be extracted from the simulator, we introduced an improved cross-entropy estimator. This improved cross-entropy estimator is used to define two new likelihood-free inference techniques: \alice and \alices. 

Our experiments comparing \alice and \alices with the other recently developed techniques indicate that they are significantly more sample efficient than \rolr, \cascal, and \rascal techniques. We attribute this to the lower variance of the improved cross-entropy estimator. For smaller training sample sizes, there are still advantages to the \sally and \sallino techniques.

We note that it is possible to use a hybrid of the traditional cross-entropy of Eq.~\ref{eq:xe_normal} and the improved cross-entropy Eq.~\ref{eq:improved_xe}. This would be useful in situations where one may not have access to the joint ratio for practical reasons or because some training samples come from real data instead of a simulation. Furthermore, we note that the improved cross-entropy estimator of \alice and \alices can be extended from the binary setting to one where samples are generated from multiple parameters points if the joint likelihood ratio for all pairs is available. These joint likelihood ratios provide the necessary ingredient for importance sampling beyond the binary setting considered here. 

The ubiquity of simulators and other implicit models indicates there is enormous potential for likelihood-free inference techniques. The use of augmented data improves the sample efficiency of these techniques significantly, and these results motivate further study of variance reduction techniques that leverage this augmented data.

\subsection*{Acknowledgments}

JB, KC, and GL are grateful for the support of the Moore-Sloan data science environment at NYU. KC and GL were supported through the NSF grants ACI-1450310 and PHY-1505463. JP was partially supported by the Scientific and Technological Center of Valpara\'{i}so (CCTVal) under Fondecyt grant BASAL FB0821. This work was supported in part through the NYU IT High Performance Computing resources, services, and staff expertise.

\bibliography{references}

\end{document}